\documentclass[letterpaper, 10 pt, conference]{IEEEtran}
\pdfoutput=1    

\usepackage{xcolor}
\definecolor{wong-black}        {HTML}{000000}
\definecolor{wong-lightorange}  {HTML}{E69F00}
\definecolor{wong-lightblue}    {HTML}{56B4E9}
\definecolor{wong-green}        {HTML}{009E73}
\definecolor{wong-yellow}       {HTML}{F0E442}
\definecolor{wong-darkblue}     {HTML}{0072B2}
\definecolor{wong-darkorange}   {HTML}{D55E00}
\definecolor{wong-pink}         {HTML}{CC79A7}

\usepackage[utf8]{inputenc}
\usepackage[margin=0.75in]{geometry} 
\usepackage{afterpage}

\usepackage[accsupp]{axessibility}  

\usepackage{url}

\usepackage{hyperref} 

\hypersetup{
    colorlinks=true,
    citecolor=wong-green,
    linkcolor=wong-darkblue,
    filecolor=wong-pink,      
    urlcolor=wong-black,
    pdfpagemode=FullScreen,
    }

\usepackage{cite}
\usepackage{amsmath,amssymb,amsfonts}
\usepackage{adjustbox}
\usepackage{algorithmic}
\usepackage{graphicx}
\usepackage{booktabs}
\usepackage{textcomp}
\usepackage{hyperref}
\usepackage{dpfloat}
\usepackage{float}
\restylefloat{table}
\usepackage{multirow}
\usepackage{threeparttable}
\usepackage{caption}
\usepackage{amsfonts}

\def\BibTeX{{\rm B\kern-.05em{\sc i\kern-.025em b}\kern-.08em
    T\kern-.1667em\lower.7ex\hbox{E}\kern-.125emX}}
    
\newcommand\nnfootnote[1]{  
  \begin{NoHyper}
  \renewcommand\thefootnote{}\footnote{#1}%
  \addtocounter{footnote}{-1}%
  \end{NoHyper}
}

\begin{document}

\title{
Collaborative Perception Datasets in Autonomous Driving: A Survey}


\author{\IEEEauthorblockN{Melih Yazgan\IEEEauthorrefmark{2}\IEEEauthorrefmark{3}\textsuperscript{\textasteriskcentered},
Mythra Varun Akkanapragada\IEEEauthorrefmark{3}\textsuperscript{\textasteriskcentered},
and J. Marius Zöllner\IEEEauthorrefmark{2}\IEEEauthorrefmark{3}}

\IEEEauthorblockA{\IEEEauthorrefmark{2}FZI Research Center for Information Technology, Germany}
\IEEEauthorblockA{\IEEEauthorrefmark{3}Karlsruhe Institute of Technology, Germany\\}}

\maketitle

\nnfootnote{\textasteriskcentered~These authors contributed equally.} \\




\begin{abstract}
This survey offers a comprehensive examination of collaborative perception datasets in the context of Vehicle-to-Infrastructure (V2I), Vehicle-to-Vehicle (V2V), and Vehicle-to-Everything (V2X). It highlights the latest developments in large-scale benchmarks that accelerate advancements in perception tasks for autonomous vehicles. The paper systematically analyzes a variety of datasets, comparing them based on aspects such as diversity, sensor setup, quality, public availability, and their applicability to downstream tasks. It also highlights the key challenges such as domain shift, sensor setup limitations, and gaps in dataset diversity and availability. The importance of addressing privacy and security concerns in the development of datasets is emphasized, regarding data sharing and dataset creation. The conclusion underscores the necessity for comprehensive, globally accessible datasets and collaborative efforts from both technological and research communities to overcome these challenges and fully harness the potential of autonomous driving.
\end{abstract}




\section{Introduction}
\label{sec:introduction}
In the evolving landscape of Intelligent Transportation Systems (ITS), there is a significant shift toward collaborative perception, which enhances the capabilities of autonomous driving and traffic management systems. Central to this shift is the implementation of V2X communications, which includes interactions such as V2V\cite{wang2020v2vnet}, V2I\cite{bai_pillargrid_2022}, and even Vehicle-to-Pedestrian (V2P)\cite{zhang_implementation_2023}. This advanced approach substantially improves traditional single-vehicle detection systems, offering a more comprehensive and accurate understanding of complex traffic environments.
One of the key advantages of collaborative perception lies in its ability to overcome the inherent limitations of individual vehicle systems, particularly in dealing with occlusions and detecting long-range objects and sensor noise\cite{li_v2x-sim_2022}. Integrating data from multiple sources increases the field of view, leading towards a holistic view of the surroundings.
This multi-faceted perception enhances safety by providing a more accurate representation of the environment and contributes to more efficient traffic flow and better decision-making capabilities for autonomous vehicles.
Established single-vehicle datasets such as KITTI\cite{geiger_are_2012}, nuScenes\cite{caesar_nuscenes_2020}, and Waymo\cite{sun_scalability_2020} do not address the complexity of collaborative perception in addition to limitations such as sensor heterogeneity, communication protocols testing, information fusion, testing and validation of collaborative perception frameworks. Recognizing these limitations, researchers have published various datasets to test and benchmark frameworks under conditions that mimic real-world scenarios involving multiple vehicles and infrastructure components.

To the best of our knowledge, this work presents the most comprehensive collection of datasets for V2V and V2I research to date, incorporating road intersection datasets. Intersections represent some of the most complex and dynamic urban traffic environments, where various agents such as vehicles, pedestrians, and cyclists interact\cite{wang2022ips300+}. For autonomous vehicles, navigating through intersections poses a formidable challenge. The unpredictability and diversity of scenarios encountered at these junctions necessitate advanced perception and decision-making capabilities\cite{ye_rope3d_2022}. Furthermore, infrastructure sensors crucially enhance perception by providing vital environmental data less prone to the blind spots and occlusions typical of vehicle-mounted sensors\cite{xu_v2x-vit_2022}. The paper's main contributions include:
\vspace{-1pt}
 \begin{itemize}
  \item Comparison between different datasets based on diversity, sensor setup, quality, public availability, and downstream tasks such as 3D object detection, object tracking, motion prediction, trajectory prediction, and domain adaptation.
  \item Comprehensive discussion on the challenges and domain gaps encountered by datasets, along with exploring the scope of future work to address these issues, is a vital aspect of this study. 
\end{itemize}
Some datasets have been excluded due to their limited size\cite{yuan2021comap}, lack of information regarding size, annotation, and benchmark \cite{busch_lumpi_2022}. 

The paper is organized as follows: Section \ref{sec:road_intersection_datasets} systematically analyzes the road intersection datasets, presenting a comparison in Table \ref{tab:road-intersection-datasets}. This is followed by a systematic analysis of collaborative perception datasets in Section \ref{sec:collaborative_perception_datasets} with a summarized comparison in Table \ref{tab:cooperative-perception-datasets-table}. Section \ref{sec:discussion} discusses open challenges for future research. Section \ref{sec:conclusion} summarizes the key findings of this review.
\begin{table*}
\caption{Overview of Road Intersection Datasets}
\label{tab:road-intersection-datasets}
\centering
\small 
\begin{threeparttable}
\begin{tabular}{p{3cm}p{1.cm}p{1.0cm}p{3cm}p{1.5cm}p{1cm}p{1.2cm}p{1.5cm}p{0.8cm}}
\toprule
\textbf{Dataset} & \textbf{Year} & \textbf{Source} & \textbf{Sensors-(Size)} & \textbf{3D Labels} &\textbf{Classes}& \textbf{Tasks} & \textbf{Website}& \textbf{Public} \\
\toprule
\textbf{BAAI-VANJEE}\cite{yongqiang_baai-vanjee_2021} & 2021 & Real & C-(2,500), L-(5,000) & 74,000 & 12& OD & \href{https://data.baai.ac.cn/details/RoadsideDataset}{Link}&\checkmark \\
\textbf{IPS300+}\cite{wang2022ips300+} & 2021 & Real & C-(14,198), L-(14,198) & 4,5M & 7& OD & \href{http://www.openmpd.com/column/IPS300}{Link}&\checkmark \\
\textbf{Rope3D}\cite{ye_rope3d_2022} & 2022 & Real & C-(50,000)& 1,5M & 13& OD & 
\href{https://thudair.baai.ac.cn/rope}{Link}&- \\
\textbf{TUMTraf-I}\cite{zimmer2023tumtraf} & 2023 & Real& C-(4,800), L-(4,800) & 57,406 & 10& OD & \href{https://innovation-mobility.com/en/project-providentia/a9-dataset/}{Link}&\checkmark\\
\textbf{RCooper}\cite{hao2024rcooper} & 2024 & Real& C-(50,000), L-(30,000) & 30,000 & 10& OD, OT & \href{https://github.com/AIR-THU/DAIR-RCooper}{Link}&\checkmark\\
\bottomrule
\end{tabular}
\begin{tablenotes}[flushleft]
    \scriptsize
    \item[1] Sensors: Camera (C), Lidar (L)
    \item[2] Tasks: Object Detection (OD), Object Tracking (OT)
\end{tablenotes}
\end{threeparttable}
\end{table*}
\section{Datasets}
\label{sec:datasets}

\subsection{Road Intersection Datasets}
\label{sec:road_intersection_datasets}   
Road intersection datasets are crucial, where the dataset requires diverse camera angles to capture the complex intersection traffic, and environment conditions present unique challenges\cite{ye_rope3d_2022}. These datasets are instrumental in refining 3D object detection and localization, addressing occlusions and truncations. Table \ref{tab:road-intersection-datasets} provides the details of the datasets at a glance for further reference.

\textbf{BAAI-VANJEE}\cite{yongqiang_baai-vanjee_2021} is an infrastructure-side open-sourced dataset with highly diverse scenes.
The data, which includes 2,500 frames of LiDAR data and 5,000 frames of RGB images, were recorded in sunny, cloudy, and rainy weather conditions. Its 74,000 3D and 105,000 2D object annotations distinguish the dataset. 
The dataset focuses on 12 classes, such as pedestrians, bicycles, motorcycles, and several types of vehicles and roadblocks.
The VANJEE smart base station, approximately 4,5 meters high at a Chinese intersection, collects the data. It is equipped with a 32-channel LiDAR sensor and four cameras.
Focusing mainly on tricycles and frames with densely packed object instances, this dataset features a significantly higher number of annotations per frame compared to the KITTI\cite{geiger_are_2012} dataset. The dataset is available under a non-commercial license.

\textbf{IPS300+}\cite{wang2022ips300+} is multi-modal, the first large-scale open-sourced roadside perception dataset. IPS300+ covers an extensive area of 3000 square meters and extends to 300 meters. Being designed from evening rush hour scenarios in the Haidian district, Beijing, China, comprises 14,198 frames. Each frame contains an average of 319,84 labels, which is significantly higher than many existing datasets like KITTI\cite{geiger_are_2012}. Labeling incorporates LiDAR point clouds and images, ensuring accurate 3D bounding box annotations for categories, including pedestrians, cyclists, tricycles, cars, buses, trucks, and engineering vehicles. Each intersection in the dataset has an 80-channel LiDAR, two RGB cameras, and one GPS, providing a comprehensive view of the surrounding environment. 
The dataset provides a label document consistent with KITTI\cite{geiger_are_2012} and contains more pedestrians and vehicles per frame than KITTI\cite{geiger_are_2012} or nuScenes\cite{caesar_nuscenes_2020}. 
The statistics in the dataset show that the annotation size of buses and trucks is relatively smaller
than that of cars, which can affect the detection accuracy of those classes. The dataset employs time synchronization and spatial calibration between different units, ensuring the consistency of labeling and accuracy of the collected multi-modal data. While the dataset's performance in 3D LiDAR detection assessed using baseline PointPillar\cite{lang_pointpillars_2019}, its evaluation with camera-based methods remains unexplored, emphasizing a notable gap in monocular approach assessments. The dataset is publicly available under the CC BY-NC-SA 4.0 license.
 
\textbf{Rope3D}\cite{ye_rope3d_2022} is another dataset specifically developed for monocular 3D object detection tasks. Rope3D includes a collection of 50,000 images and 1.5 million 3D annotations. Comprising 13 object classes, it provides a detailed representation of roadside elements. The primary classes include cars, oversized vehicles, pedestrians, cyclists, and extra classes such as traffic cones, triangle plates, and unknown-unmovable objects. The dataset utilizes roadside cameras mounted on poles or traffic lights and LiDAR sensors equipped on vehicles parked or moving. Rope3D comprises images captured across various lighting conditions, weather conditions, and diverse road scenes. Each has distinct camera specifications, including focal length, pitch angle, and mounted height. The annotation process focused on accurately aligning 2D and 3D annotations and paying particular attention to occlusion and truncation levels. 
This approach utilizes a model to bridge 3D locations and 2D projections, effectively addressing the challenge of non-parallel optical axes in cameras with varying pitch angles, thereby solving the gap as previously discussed in IPS300+\cite{wang2022ips300+}.
Benchmarking involves evaluating adapted monocular 3D object detection such as  M3D-RPN\cite{brazil_m3d-rpn_2019}. Addressing ethical concerns, Rope3D anonymizes sensitive information such as license plates and human faces. It also restricts the use of time-discrete images to prevent potential misuse for illegal surveillance. The dataset is governed by strict usage terms outlined in a detailed confidentiality and use agreement, restricting its availability outside specified conditions and prohibiting open-source distribution.

\begin{figure}[h]
\center
\includegraphics[width=8cm, height=4cm]{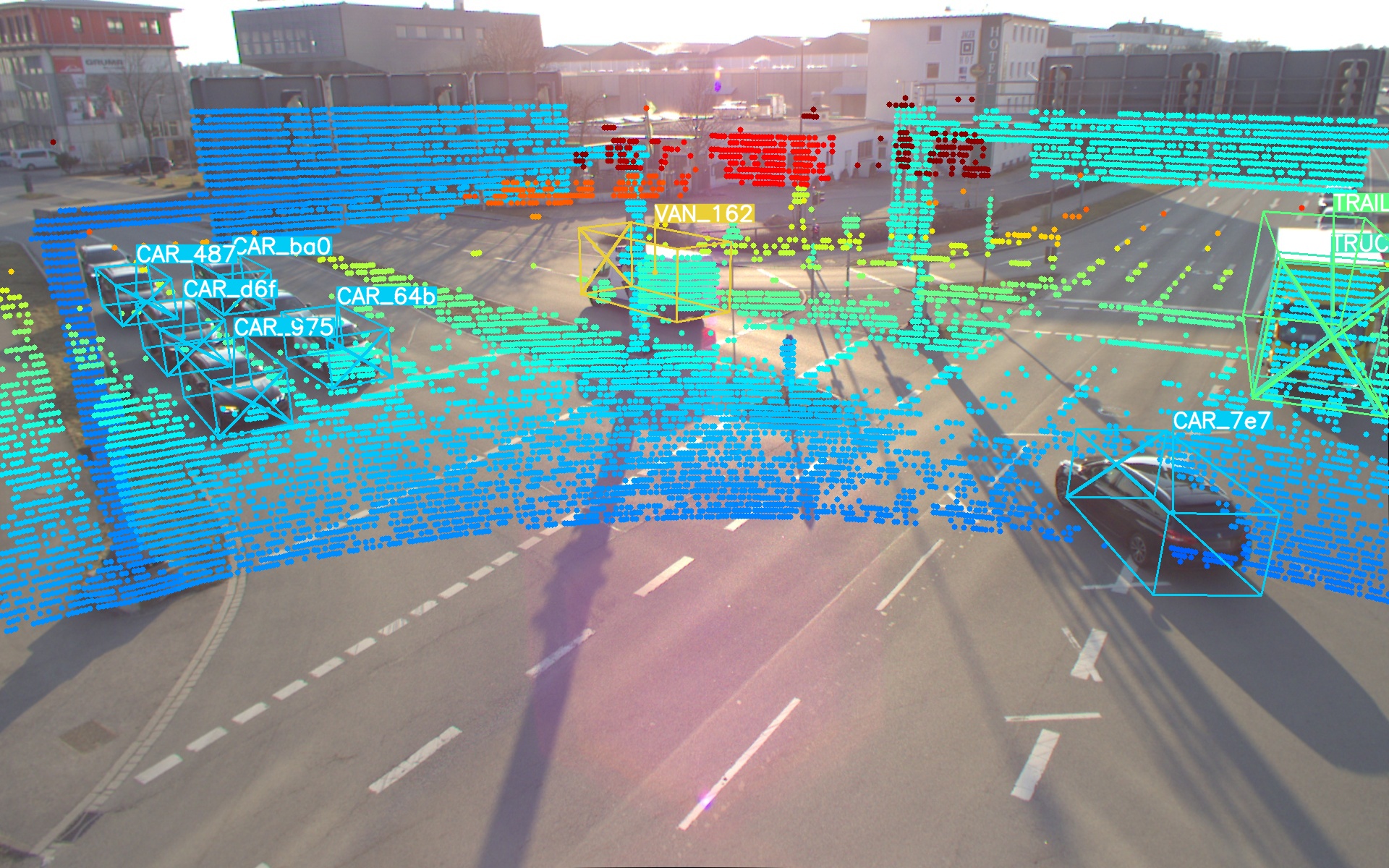}
\caption{3D labels with LiDAR points on camera frame\cite{zimmer2023tumtraf}.}
\label{fig:TUMTraf_pic}
\centering
\end{figure}

\textbf{TUMTraf-I}\cite{zimmer2023tumtraf} is another multi-modal large-scale open-sourced dataset. 
TUMTraf-I comprises 4,800 images and LiDAR point cloud frames, which include over 57,406 labeled 3D annotations. These are partitioned into ten distinct object classes of traffic participants, offering a wide range of classes in real-world scenarios. The classes include cars, trucks, trailers, vans, pedestrians, motorcycles, buses, bicycles, emergency vehicles, and others. 
This dataset is equipped with two cameras and two LiDARs mounted at a height of 7 meters, offering a 360° field of view. This elevated perspective is essential for observing traffic scenarios, such as turns, overtaking maneuvers, and lane merges.
Accordingly, the dataset is segmented into four subsets (S1 to S4), each encapsulating different atmospheric conditions and providing a realistic spectrum of driving scenarios. Subsets S1 and S2 capture 30-second sequences during dusk, presenting continuous camera footage alongside labeled LiDAR captures. Conversely, subset S3 offers a detailed 120-second sequence shot in bright daylight, while subset S4 provides a 30-second nighttime recording amidst heavy rainfall.
TUMTraf-I is used to assess 3D object detection, employing PointPillars\cite{lang_pointpillars_2019} for LiDAR-based and MonoDet3D\cite{zimmer2023infra} for camera-based approaches. Additionally, TUMTraf-I provides a development kit. The kit supports multiple dataset formats, offering versatility and ease of integration with existing models and systems. Fig \ref{fig:TUMTraf_pic} is created with the development kit. 
The dataset is publicly available under the License CC BY-NC-ND 4.0.

\textbf{RCooper}\cite{hao2024rcooper} is the latest real-world dataset specifically developed for roadside cooperative perception tasks. RCooper includes 50,000 images and 30,000 point clouds, covering two primary traffic scenes: intersections and corridors. This sets it apart from other datasets focused solely on intersections. RCooper provides ten object classes, which include various vehicles, cyclists, pedestrians, and construction elements.
Each scene features tailored sensor setups to address specific topological challenges: intersections are equipped with a combination of MEMS LiDARs and 80-32 channel LiDARs, both operating at 10Hz, along with cameras, to capture the dynamic and congested nature of urban crossroads adeptly. Corridors are monitored with similar LiDARs and cameras, ensuring extensive coverage along extended road stretches. This varied deployment of sensors, particularly the distinction in LiDAR technologies, significantly enhances the dataset's utility for exploring challenges related to sensor heterogeneity. The benchmarking for RCooper includes evaluating cooperative perception tasks like 3D object detection and tracking using state-of-the-art methods such as \cite{chen_f-cooper_2019,xu2022opencood,xu_cobevt_2022}. The dataset is publicly available.

\subsection{Collaborative Perception Datasets}
\label{sec:collaborative_perception_datasets} 
The collaborative perception is witnessing significant advancements through the development of datasets. These datasets focus on enhancing V2V and V2X communication. By simulating complex urban environments and diverse driving scenarios, they contribute to developing algorithms for various tasks. The intricate data provided by these datasets, including detailed frame collections and comprehensive sensor setups as demonstrated in Table \ref{tab:cooperative-perception-datasets-table}, are pivotal in addressing the challenges of dynamic road conditions. For further details, reference to the official dataset pages, linked within the dataset names in Table \ref{tab:cooperative-perception-datasets-table}, is encouraged.

\begin{table*}
\caption{Overview of Collaborative Perception Datasets}
\label{tab:cooperative-perception-datasets-table}
\centering
\small
\begin{threeparttable}
\begin{tabular}{p{2.7cm}p{0.6cm}p{0.8cm}p{1.5cm}p{1.0cm}p{0.8cm}p{0.8cm}p{2.7cm}p{1cm}}
\toprule
\textbf{Dataset} & \textbf{Year} & \textbf{Source} &\textbf{V2X} &\textbf{Sensors} & \textbf{Size}& \textbf{Agents} & \textbf{Tasks}  & \textbf{PA/R}\\
\toprule
\href{https://ai4ce.github.io/V2X-Sim/download.html}{\textbf{V2X-Sim 1.0}}\cite{li_learning_2022} & 2022 & Sim & V2V & L & 10,000 &2-5 &OD, OT, SS &\checkmark/\checkmark\\
\href{https://ai4ce.github.io/V2X-Sim/.}{\textbf{V2X-Sim 2.0}}\cite{li_v2x-sim_2022} & 2022 & Sim & V2V, V2I& C, L & 10,000 & 2-5& OD, OT, SS &\checkmark/\checkmark\\
\href{https://mobility-lab.seas.ucla.edu/opv2v/}{\textbf{OPV2V}}\cite{xu_opencdaopen_2021} & 2022 & Sim& V2V &C, L & 11,464 & 2-7 & OD, OT &\checkmark/\checkmark \\
\href{https://github.com/AIR-THU/DAIR-V2X.}{\textbf{DAIR-V2X-C}}\cite{yu_dair-v2x_2022} & 2022 & Real & V2I& C, L & 38,845 & 2 & OD &- / -\\
\href{https://github.com/DerrickXuNu/v2x-vit} {\textbf{V2XSet}}\cite{xu_v2x-vit_2022} & 2022 & Sim & V2V, V2I& C, L & 11,447 & 2-7 & OD&\checkmark/ -\\
\href{www.dolphins-dataset.net.}{\textbf{DOLPHINS}}\cite{mao2022dolphins}  & 2023 & Sim & V2V, V2I& C, L & 42,736 & 3 & OD&\checkmark/\checkmark\\
\href{https://doi.org/10.25835/75o9yrc0}{\textbf{LUCOOP}}\cite{axmann_lucoop_2023} & 2023 & Real & V2V & L & 54,000 & 3 &OD, OT &\checkmark/ -\\
\href{https://mobility-lab.seas.ucla.edu/v2v4real/}{\textbf{V2V4Real}}\cite{xu_v2v4real_2023} & 2023 & Real & V2V & C, L & 60,000 & 2 &OD, OT, DA &\checkmark/ -\\
\href{https://github.com/AIR-THU/DAIR-V2X-Seq.}{\textbf{V2X-Seq(SPD)}}\cite{yu_v2x-seq_2023} & 2023 & Real & V2I& C, L & 15,000 &2& OD, OT, TP &- / - \\
\href{https://deepaccident.github.io/}{\textbf{DeepAccident}}\cite{wang_deepaccident_2023} & 2023 & Sim & V2V, V2I & C, L & 57,000 & 1-5 & OD, OT, SS, MP, DA &\checkmark/ -\\
\href{https://tum-traffic-dataset.github.io/tumtraf-v2x/}{\textbf{TumTraf-V2X}}\cite{zimmer2024tumtraf} & 2024 & Real & V2I & C, L & 7,500 & 2 & OD, OT &\checkmark/ -\\
\bottomrule
\end{tabular}
\begin{tablenotes}[flushleft]
    \scriptsize
    \item[1] Public Availability (PA), Reusability (R)
    \item[2] Sensors: Camera (C), Lidar (L)
    \item[3] Tasks: Object Detection (OD), Object Tracking (OT), Semantic Segmentation (SS), Trajectory Prediction (TP), Motion Prediction (MP), Domain Adaptation (DA)
\end{tablenotes}
\end{threeparttable}
\end{table*}


\textbf{V2X-Sim 1.0}\cite{li_learning_2022} is another open-source V2V dataset, even though the dataset's name suggests otherwise. The dataset is created by using CARLA\cite{dosovitskiy2017carla} and SUMO\cite{krajzewicz2012recent}. 
It features a 32-channel LiDAR system with a 70-meter range, operating in a dense traffic simulation within the Town05 environment.
Each scenario includes 20s  traffic flow and recordings at 5Hz.
Within each scene, 2-5 vehicles are randomly chosen as Connected Autonomous Vehicles (CAVs). 
The dataset format, derived from nuScenes\cite{caesar_nuscenes_2020}, is extended to multi-agent scenarios, containing 10k frames in total.
The dataset focuses on the 3D perception task and proposes a trainable, dynamic collaboration graph to control agent communication.
Comprehensive benchmarks conducted in 3D object detection have shown that the proposed DiscoNet\cite{li_learning_2022} outperforms methods such as V2VNet\cite{wang2020v2vnet}, Who2com\cite{liu_who2com_2020}, and When2com\cite{liu_when2com_2020} in terms of the performance-bandwidth trade-off and communication latency.
The dataset has been designed to be reproducible for future research.

\begin{figure*}
\centering
\includegraphics[width=10cm, height=7.5cm]{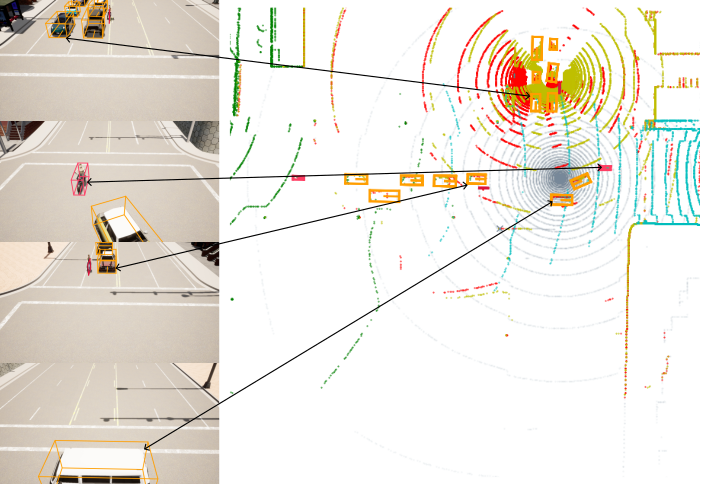}
\caption{The left panel shows the RSU detection frames and the right panel illustrates a LiDAR point cloud dataset, where the RSU is denoted in grey and an array of distinct colors distinguishes the various CAVs\cite{li_v2x-sim_2022}.}
\label{fig:V2XSim_pic}
\end{figure*}

\textbf{V2X-Sim 2.0}\cite{li_v2x-sim_2022} is the first open-source simulated V2X dataset, a V2I extension version of V2X-Sim 1.0\cite{li_learning_2022}. It captures traffic flow at intersections in three different CARLA towns, maintaining the same frame rate and total frame count as V2X-Sim 1.0\cite{li_learning_2022}.
Each vehicle includes RGB cameras, LiDAR, GPS, and IMU, while Road Side Units (RSUs) are outfitted with RGB cameras and LiDAR as demonstrated in Fig \ref{fig:V2XSim_pic}. 
Vehicles have six RGB cameras based on the nuScenes\cite{caesar_nuscenes_2020} configuration, and RSUs have four cameras pointing in all directions at intersections.
The dataset is benchmarked simultaneously for 3D BEV object detection, tracking, and segmentation with intermediate collaborative methods\cite{wang2020v2vnet,liu_when2com_2020,liu_who2com_2020,li_learning_2022}. The dataset is available under a non-commercial
license. The entire dataset with V2X-Sim 1.0\cite{li_learning_2022} is open-sourced.

\textbf{OPV2V}\cite{xu2022opencood} is another simulated open-source dataset for V2V communication, includes various roadway types and scenarios.
The dataset was generated using CARLA\cite{dosovitskiy2017carla} in conjunction with the OpenCDA\cite{xu_opencdaopen_2021} co-simulation tool, featuring multiple CAVs equipped with a comprehensive sensor setup. It comprises more than 70 scenes, 11,464 frames, and 232,913 annotated 3D vehicle bounding boxes, gathered from eight towns in CARLA and the digital town of Culver City, Los Angeles. Each frame has, on average, approximately three CAVs, with a minimum of two and a maximum of seven CAVs. Each CAV has four cameras, 64-channel LiDAR, and GPS/IMU sensors. The sensor data is streamed at 20 Hz and recorded at 10 Hz. The dataset covers frames from short scenarios in six road types: suburban midblock, urban T-intersection, urban curved road, freeway entrance ramp, urban 4-way intersection, and rural curvy road.
The dataset supports collaborative 3D vehicle detection, BEV semantic segmentation, and tracking tasks only in V2V scenarios. Its benchmarking includes three fusion strategies, with the effect of CAV quantity and detection accuracy-compression trade-off. These are applied only in 3D Lidar-based object detection methods like VoxelNet\cite{zhou_voxelnet_2018} and PointPillar\cite{lang_pointpillars_2019}.
The dataset is made fully reproducible through the inclusion of driving logs. The dataset is available under a non-commercial license.

\textbf{DAIR-V2X}\cite{yu_dair-v2x_2022}
is a large-scale V2I collaborative perception dataset derived from the real world. It features 71,254 LiDAR and camera frames with various vehicle types and pedestrians, including cyclists and motorcyclists. The dataset encompasses various environments, including 10 kilometers of urban roadways, an equal distance on highways, and 28 distinct intersections, all captured under varying weather conditions and lighting scenarios.
The dataset is divided into three main subsets, with the DAIR-V2X-C subset focusing on V2I collaboration. This subset is particularly notable for introducing the Time Compensation Late Fusion (TCLF) framework, which was developed to address the challenges of temporal asynchrony by using a specialized asynchronous subset from DAIR-V2X-C. 
Alongside the DAIR-V2X-C subset, the dataset also features the DAIR-V2X-V and DAIR-V2X-I subsets, focusing on vehicle and infrastructure only.
Unlike others, the dataset incorporates both 3D LiDAR and image detection. For LiDAR detection, it leverages  PointPillar\cite{lang_pointpillars_2019} and implements both early and late fusion techniques, accommodating synchronous and asynchronous data, and includes the TCLF framework. The late fusion framework utilizes ImvoxelNet\cite{rukhovich2022imvoxelnet} as the 3D detector with synchronous data for image detection. The license conditions of this dataset mirror those of Rope3D\cite{ye_rope3d_2022}, adhering to identical usage and distribution terms.

\textbf{V2XSet}\cite{xu_v2x-vit_2022}
is an open-source simulation dataset that considers real-world challenges in V2X collaboration using CARLA\cite{dosovitskiy2017carla} and OpenCDA\cite{xu_opencdaopen_2021}. It comprises 55 representative scenes covering five roadway types: straight, curvy, intersection, midblock, and entrance from eight towns. Statistical analysis shows that the dataset is biased on intersection data. Comprising 11k frames, V2XSet incorporates both V2X cooperation and realistic noise simulation, unlike DAIR-V2X\cite{yu_dair-v2x_2022} or OPV2V\cite{xu2022opencood}.
Each vehicle is equipped with 32-channel LiDAR mounted on the top and infrastructure sensors at approximately 4.5 meters, which record at 10 Hz. Each scene contains at least two and, at most, seven intelligent agents and lasts 25 seconds.
The dataset is used for evaluating the effect of spatial and temporal uncertainties on 3D object detection accuracy in collaborative intermediate fusion methods such as V2VNet\cite{wang2020v2vnet}, AttFuse\cite{xu2022opencood}, F-Cooper\cite{chen_f-cooper_2019}, and DiscoNet\cite{li_learning_2022}. The dataset is released under a non-commercial license.

\textbf{DOLPHINS}\cite{mao2022dolphins}
is a large-scale, open-source V2X dataset generated using CARLA\cite{dosovitskiy2017carla}. 
It distinguishes itself from other simulation datasets by featuring dynamic weather conditions across 42,736 frames and 292,549 3D annotations compatible with the KITTI format\cite{geiger_are_2012}.
The dataset includes at least three agents per scenario, each equipped with 64-channel LiDAR and RGB cameras, providing synchronized images and point clouds from CAVs and RSUs. 
DOLPHINS covers six autonomous driving scenarios: urban intersections, T-junctions, steep ramps, highways on-ramps, and uniquely mountain roads and lane merging, unlike V2X-Sim\cite{li_v2x-sim_2022} datasets, which have a limited viewpoint on specific scenarios.
In addition to standard labeling, the dataset is enriched with two key information types: the positions of surrounding vehicles and context-sensitive labels. These elements are vital for synchronizing perception data from various viewpoints. They cover all traffic entities within a 100-meter radius ahead and behind the ego vehicle and 40 meters on either side, providing extensive and detailed coverage.
The dataset focuses on vehicle and pedestrian detection and supports 2D and 3D object detection in single-vehicle Perception. Further, it benchmarks the early fusion LiDAR 3D object detection with PointPillars\cite{lang_pointpillars_2019} and MVX-Net\cite{sindagi_mvx-net_2019}. Along with the dataset, the corresponding codes are released for flexibility and extendability of the dataset on-demand. The dataset is released under a CC BY-NC-SA 4.0 license.

\textbf{LUCOOP}\cite{axmann_lucoop_2023} is a large scale real-world V2V dataset created by Leibniz University. It stands out from the other real-world datasets, focusing on multi-vehicle urban navigation and collaborative perception. The LUCOOP dataset encompasses over 54,000 LiDAR frames, approximately 700,000 IMU measurements, 3D map point clouds, and more than 2.5 hours of 10 Hz GNSS raw data.
The dataset is gathered from three vehicles equipped with LiDAR, GNSS, IMUs, and Ultra-Wide-Band (UWB) sensors, capturing a detailed view of urban environments with narrow streets and tall buildings. Furthermore, it is enriched with a LOD2\cite{LGLN2022} city model, enhancing its urban simulation capabilities.
Integrating a stationary total station and static UWB sensors is crucial for improving the dataset's accuracy. This integration contributes over 6,000 high-precision measurements that cover more than 1 km of the vehicle's trajectory. This level of granularity and precision in ground truth verification is particularly valuable for V2V and V2X range measurements. 
The dataset provides further 3D bounding box annotations and precise vehicle poses but includes no benchmarking. The dataset is published with a CC BY-NC 3.0 License.

\begin{figure}
\centering
\includegraphics[width=8cm, height=4cm]{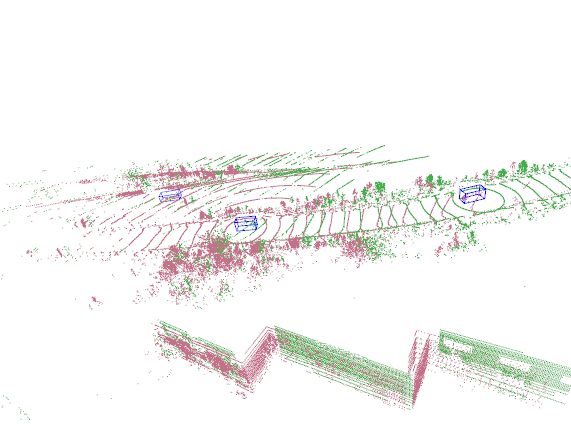}
\caption{Lidar point clouds, coloring relative to agents\cite{xu_v2v4real_2023}.}
\centering
\label{fig:V2V4Real_pic}
\end{figure}

\textbf{V2V4Real}\cite{xu_v2v4real_2023} is another large-scale, real-world, multi-model dataset for V2V perception. The dataset is collected in Columbus, Ohio, and features a diverse sensor suite on two vehicles. It has 240,000 annotated 3D bounding boxes and uniquely HDMaps across five vehicle classes captured over diverse road types, including intersections, highway ramps, and urban roads. 
Equipped with LiDAR, front and rear mono cameras, and GPS/IMU systems, the dataset ensures comprehensive data capture at 10Hz. 
The vehicles covered 410 km of road, maintaining a distance within 150 meters to guarantee overlapping sensor views as demonstrated in Fig \ref{fig:V2V4Real_pic}.
To address potential overlaps in object identification, each vehicle in the dataset is assigned a unique range of object IDs, ensuring clear differentiation.
V2V4Real's benchmarking includes three fusion strategies: Late Fusion, Early Fusion, and leading intermediate methods such as AttFuse\cite{xu_opencdaopen_2021}, F-Cooper\cite{chen_f-cooper_2019}, V2VNet\cite{wang2020v2vnet}, V2XVit\cite{xu_v2x-vit_2022}, and CoBEVT\cite{xu_cobevt_2022}. These are applied over three cooperative perception tasks: 3D object detection, object tracking, and sim-to-real domain adaptation. Lidar data in OPV2V\cite{xu2022opencood} and KITTI\cite{geiger_are_2012} format is available for download. The dataset is available under a non-commercial license.

\textbf{V2X-Seq}\cite{yu_v2x-seq_2023} is the first large-scale sequential dataset, offering data collected from real-world scenarios. Unlike DAIR-V2X\cite{yu_dair-v2x_2022}, which focuses on 3D object detection, V2X-Seq is uniquely designed for tracking and trajectory forecasting tasks. It consists of two main parts: the Sequential Perception Dataset (SPD) and the Trajectory Forecasting Dataset (TFD). SPD, an extension of DAIR-V2X-C\cite{yu_dair-v2x_2022}, includes over 15,000 frames from 95 scenarios, each lasting 10–20 seconds. It features vehicle and infrastructure frames sampled at 10Hz, equipped with 3D annotations for ten object classes, including unique tracking IDs for each object. TFD, on the other hand, comprises about 80,000 infrastructure-view, 50,000 vehicle-view, and 50,000 cooperative-view scenarios from 28 intersections. This subset covers 672 hours of data, providing sequences of tracked object data for 10 seconds. Additionally, the dataset includes real-time traffic light signals recorded at 10 Hz for the infrastructure portion of TFD. This data encompasses the timestamp, location, color status, shape status, and remaining time, offering significant insights into traffic participant behaviors and interactions. The V2X-seq dataset addresses challenges related to latency, and the proposed FF-Tracking method tackles the tracking task.
Besides, V2X-Seq provides vector maps for intersection areas, organized similarly to Argoverse\cite{chang_argoverse_2019}. These maps contain detailed representations of lane centerlines, crosswalks, stop lines, and essential attributes like lane width and turn directions. These are crucial for building spatial context in trajectory analysis. The license conditions of this dataset mirror those of Rope3D\cite{ye_rope3d_2022}, adhering to identical usage and distribution terms.

\textbf{DeepAccident}\cite{wang_deepaccident_2023} is another large-scale open-source V2X dataset generated with CARLA\cite{dosovitskiy2017carla} to represent diverse accident scenarios. It is the first simulated dataset that supports a motion prediction task. Compared to the V2X-Seq\cite{yu_v2x-seq_2023}, it doesn't rely on precise vehicle locations, map topology, and traffic light information. The dataset features 57,000 annotated frames recorded at 10 Hz. It encompasses a variety of scenarios, including different road types, weather conditions, and times of day. 
The dataset's unique creation involved capturing scenes with two vehicles having overlapping planned trajectories. Additionally, two vehicles following each accident-involved vehicle and one infrastructure unit facing the intersection, summing up to five agents. Each agent has six RGB cameras and one 32-channel LiDAR.
The dataset classes consist of vehicle types, including motorcycle, cyclist, and pedestrian. 
Specifically, DeepAccident concentrates on twelve varieties of accident scenarios at intersections, including those with and without traffic control signals.
These scenarios range from running against a red light at four-way intersections to unprotected left turns and conflicting turns at three-way intersections. 
Besides its main focus on end-to-end motion and accident prediction, the dataset supports 3D object detection, tracking, and BEV semantic segmentation. Regarding benchmarking, the dataset's baseline model, V2XFormer, is compared against various state-of-the-art intermediate fusion modules. These include DiscoNet\cite{li_learning_2022}, V2X-ViT\cite{xu_v2x-vit_2022}, and CoBEVT\cite{xu_cobevt_2022}. Finally, real-world applicability tests using the nuScenes\cite{caesar_nuscenes_2020} dataset reveal improved performance with models trained on both DeepAccident and nuScenes\cite{caesar_nuscenes_2020} data. The license conditions of the dataset are unknown, but the dataset is open-sourced.

\textbf{TumTraf-V2X} \cite{zimmer2024tumtraf} dataset, derived from real-world data, is the latest to be released as open-source. It includes 2,000 labeled point clouds and 5,000 images, with approximately 30,000 3D bounding boxes that are enhanced with precise GPS and IMU data for accurate object location and movement tracking. Annotations conform to the ASAM OpenLABEL\cite{asamOpenLABEL2023} format, and the dataset features a heterogeneous sensor setup: 32-64 channel LiDARs operating at 10 Hz and high-resolution cameras. It records a broad spectrum of traffic scenarios under various environmental conditions, including complex maneuvers such as overtaking and U-turns, and instances of traffic violations, setting it apart as a unique resource among real V2X datasets.
Central to this dataset is the CoopDet3D model, a V2X cooperative perception model that utilizes vehicle and infrastructure data to improve object detection and tracking. The accompanying TUMTraf V2X development kit facilitates this data collection, providing data processing, visualization, and evaluation tools. The entire package is available under a CC BY-NC-SA 4.0 license.
\section{Discussion}
\label{sec:discussion}
By comprehensively reviewing 16 collaborative perception datasets, we have identified critical areas such as domain shift, sensor setup limitations, dataset diversity, and availability. Addressing these concerns is essential for accelerating the development of autonomous driving technologies.

\textbf{Domain Shift:} According to Table \ref{tab:cooperative-perception-datasets-table}, it is clear that most of the datasets are created by using simulated environments, and only two of them evaluated the datasets with domain adaptation techniques. Due to inherent challenges such as labeling, privacy, and investment in gathering comprehensive real-world datasets, domain shift will likely remain an issue soon. As a result, the reliance on simulated datasets and the subsequent need for effective domain adaptation techniques are expected to be ongoing areas of focus in developing collaborative perception systems.

\textbf{Sensor Setup and Limitations:} As presented in Table \ref{tab:road-intersection-datasets}, the datasets are created using multiple sensor modalities. However, a critical observation from the datasets listed in Table \ref{tab:cooperative-perception-datasets-table} is the inconsistency in multi-modal approaches, especially in real-world scenarios. This indicates a significant gap in capturing the diverse and complex real-world driving conditions. Addressing limitations in vehicle and infrastructure sensors, especially under changing weather and varying light conditions, is a crucial area for further research and development. Challenges such as dealing with diverse camera angles, handling occlusions, and overcoming depth perception issues from various viewpoints are essential to address. These issues are critical for ensuring the effectiveness and reliability of data collected for both V2V and V2I applications. For V2I applications, in particular, the strategic choice of sensor heights and types necessitates further exploration. Optimizing sensor placement is key to enhancing data capture quality, which is fundamental for accurate and comprehensive environment perception.
Furthermore, the future of autonomous driving, where all cars are interconnected, introduces a new layer of complexity due to the heterogeneity of sensor modalities. Car manufacturers may employ varied sensor setups, leading to diverse data types and formats. This diversity necessitates the development of effective strategies for handling and interpreting these various data.

\textbf{Dataset Diversity:} The study conducted by Xiang et al.\cite{xiang_v2xp-asg_2023} provides a pivotal understanding of perception in challenging scenarios, which is essential for safe and robust collaborative perception in vehicle-to-everything communication systems. These scenarios extend beyond the typical occlusions, accident prediction in 12 scenarios. Trajectory prediction will play a crucial role, especially in environments where the number of CAVs exceeds the average (see Table \ref{tab:cooperative-perception-datasets-table}). Another essential improvement is integrating Vulnerable Road Users (VRUs) into V2X communications systems, particularly V2P. Additionally, advanced tasks like anomaly detection and out-of-distribution\cite{bogdoll_perception_2023} analysis are integral for evaluating and responding to unforeseen and potentially hazardous events.

\textbf{Dataset Availability:} The practice of open-sourcing datasets is crucial for promoting transparency and collaboration within the global research community, enabling researchers worldwide to address challenges with a shared resource pool. However, as indicated in Tables \ref{tab:road-intersection-datasets} and \ref{tab:cooperative-perception-datasets-table} of the paper, there is a noticeable disparity in the availability of simulated versus real-world datasets. While simulated datasets are generally accessible, many comprehensive real-world datasets remain restricted, particularly in certain regions. This lack of access presents a major obstacle for researchers needing real-world data to test and advance V2X applications, regardless of location.
Furthermore, the reusability of datasets, especially simulated ones, is increasingly important. Creating flexible datasets for adaptation or expansion to include specific user scenarios or tasks enhances their value and longevity. Integrating them into a unified framework like OpenCOOD\cite{xu2022opencood} simplifies their application in benchmarking and comparative studies.

\textbf{Privacy and Security:} The development of autonomous driving relies heavily on extensive data, including images and videos from onboard and exterior cameras. Including personal details like faces, dates, and locations in this data raises concerns about privacy and security, particularly when the collected data is used for tracking or monitoring individuals without their consent. This is especially important in the creation of real-world datasets. Collaboration faces significant challenges, particularly in dataset creation, due to security concerns like malicious attacks, especially regarding sharing sensor-captured data. In response to these challenges, using Federated Vehicular Transformers, proposed by Tian et al.\cite{tian_federated_2022}, offers a promising direction.
\section{Conclusion}
\label{sec:conclusion}
In conclusion, our comprehensive overview of collaborative perception datasets has highlighted key advancements along with persistent challenges that need to be addressed. Technological progress in this area is evident, but there are notable gaps in the availability of real-world V2X datasets. Addressing these gaps is crucial for the global research community to fully realize the potential of collaborative perception. Collaborative efforts between technology innovators and the research community are essential in this endeavor. The development of extensive, globally accessible datasets will play a pivotal role in overcoming these challenges and unlocking the full capabilities of autonomous vehicles. Our review not only highlights critical gaps but also outlines a pathway for future advancements. By underlining the importance of diverse, real-world datasets and improved sensor setups, our findings encourage the development of more adaptable and robust systems. We advocate for increased collaboration and innovation in dataset creation, aiming to accelerate the progress of autonomous vehicle capabilities.
\section{Acknowledgment}
\label{sec:acknowledgment}
This work is developed within the framework of the "Shuttle2X" project, funded by the Federal Ministry for Economic Affairs and Climate Action (BMWK) and the European Union, under the funding code 19S22001B. The authors are solely responsible for the content of this publication.

{\small
\bibliographystyle{IEEEtran}
\bibliography{referencens}

}

\end{document}